# From Artificial Intelligence to Brain Intelligence:
# The basis learning and memory algorithm for brain-like intelligence

**Authors:** Yifei Mao[1]*

**Affiliations:**

[1]410000 Changsha, Hunan, China.

*Corresponding author. Email: maoyifei94@163.com.

**Abstract:** The algorithm of brain learning and memory is still undetermined. The backpropagation algorithm of artificial neural networks was thought not suitable for brain cortex, and there is a lack of algorithm for memory engram. We designed a brain version of backpropagation algorithm, which are biologically plausible and could be implemented with virtual neurons to complete image classification task. An encoding algorithm that can automatically allocate engram cells is proposed, which is an algorithm implementation for memory engram theory, and could simulate how hippocampus achieve fast associative memory. The role of the LTP and LTD in the cerebellum is also explained in algorithm level. Our results proposed a method for the brain to deploy backpropagation algorithm, and sparse coding method for memory engram theory.

Recently, artificial intelligence based on deep neural networks has made great progress in games, medical and other fields (*1,2*). Backpropagation (BP) as the optimization algorithm of artificial neural network (*3*), is considered biologically implausible (*4*). The learning algorithm of biological neural network is still uncertain. The framework Neural Gradient Representation by Activity Differences (NGRAD) was proposed to be a solution for brain to implement BP algorithm, but there is currently a lack of ideal experimental results (*5*). It is still a mystery that whether the learning algorithm of the cerebral cortex is backpropagation or other algorithms .

In terms of memory, The memory engram theory could explain the formation and store of memory in the brain, which was verified by many neuroscience experiments (*6,7*). But there is a lack of algorithm to simulate memory engram. Engram cells and Long-term Potentiation (LTP) in the hippocampus are important for hippocampal quick memory (*8-11*).

The cerebellum is related to short-term motor learning. There are a large number of granular cells in the cerebellum, and the synaptic connection established with Purkinje cells can be induced LTP or Long-Term Depression (LTD) (*12*). These synaptic plasticity is the mechanism basis for the short-term motor learning of the cerebellum.

In this article, we try to solve three algorithm problem about learning and memory of brain: (i) How does brain implement backpropagation algorithm with biological neurons. (ii)The principle of memory engram cells formation and how engram cells participate in hippocampal memory. (iii) What is the role of LTP/LTD in cerebellar learning.

**Brain version of backpropagation algorithm**

The main idea about the brain version of backpropagation algorithm is to turn end-to-end supervision into layer-by-layer supervision. We designed a layer-wise backpropagation (LWBP) algorithm, which is modified from the BP algorithm. The process comparison between LWBP and standard BP for training deep network are shown as shown in Fig. 1a. BP calculates the loss only in the last layer, and its error gradient flow will pass through the entire network, shown as the green arrow line. LWBP algorithm calculates the error in each module separately, and its gradient will only propagate within one module, which is more biologically reasonable than BP. The internal structure of the LWBP module is shown in Fig. 1b. LWBP algorithm requires shortcut connection to achieve the best performance, which is the feature of residual network (*13*). Input data is processed by the nonlinear layer, and merged with shortcut connection to form the output signal. Loss layer processes the copy of output signal and compares it with Label for error calculation (see Methods for detail). The optimization algorithm within the module is backpropagation, but error gradient flow will only propagate within the module, shown as the green line in Fig. 1b. The shortcut connection is very important for LWBP algorithm. Two network that consist of 12 stacked LWBP module, with or without shortcut connection separately, are trained on cifar-10 (*14*) data with the same training epochs. Accuracy of every module are plotted in Fig. 1c. LWBP module with shortcut connection could make better use of deep structure. The training efficiency of LWBP and BP are compared as shown in Fig. 1d. The depth of the three experimental network are 12, number of neurons in each layer of three experiment are 300. Accuracy of the last module is taken as the accuracy of LWBP network. LWBP with shortcut connection takes 10 times as much epochs as BP algorithm to achieve the same training accuracy. The efficiency is lower if LWBP is without shortcut connection. But this is not a problem for brain, because brain has enough time to learn. In summary, LWBP is the algorithm that could make use of deep networks as BP, but does not require long-distance gradient propagation, which is biologically plausible. Its training effect is as good as BP, and it takes a longer time.

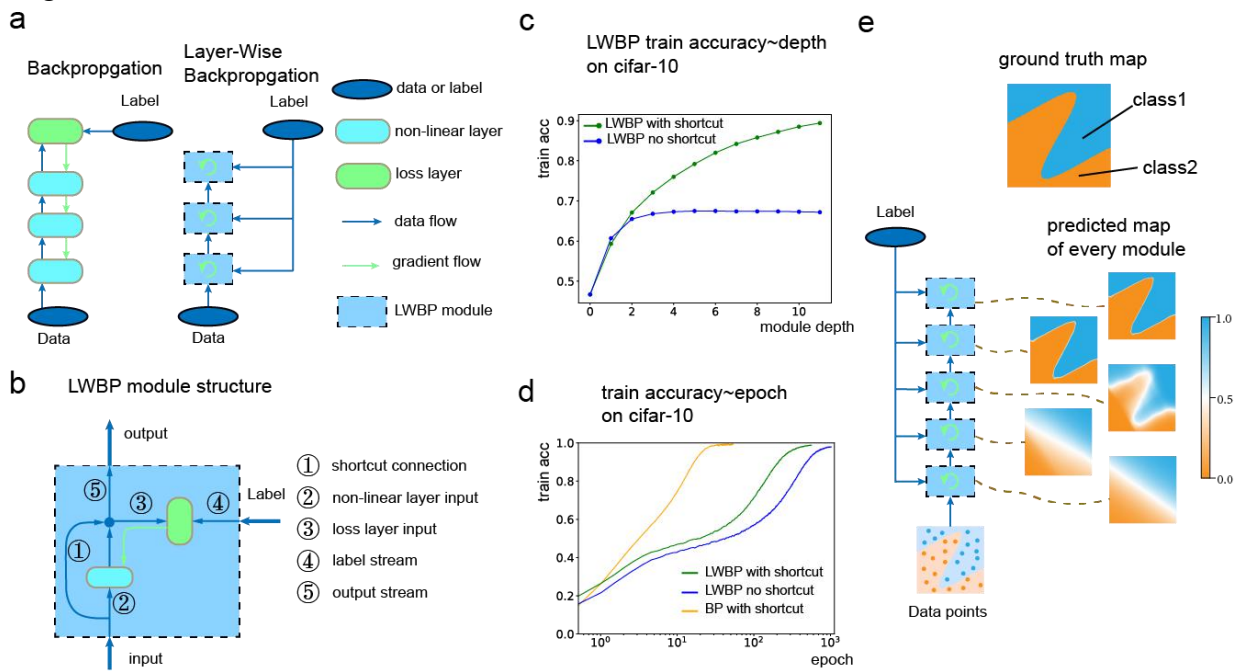

**Fig. 1| The diagram, module structure and performance of LWBP algorithm. a**, Process comparison between BP algorithm and LWBP algorithm. **b**, the internal structure of LWBP module. **c**, Layer-wise accuracy of networks that consist of 12 stacked LWBP modules, with or without shortcut connection separately. **d**, Training accuracy curves of three experiments for

comparing the efficiency of BP and LWBP. **e**, After trained on 2d point classification tasks, the prediction class map of every LWBP module .

Why the combination of LWBP and shortcut connection could train deep network to have high accuracy? According to the information bottleneck theory, the learning of a deep neural network is to retain the information related to the target and remove the information unrelated to the target (*15,16*). Each module of LWBP network retains target related information by supervised learning, and a module with shortcut connection can ensure its accuracy no worse than previous module, by degenerating into an identity mapping. We built a 2d point classification task to visualize the mechanism of LWBP. The task is a binary classification task of points in a rectangular area, and the network consists of 5 LWBP module, as shown in Fig. 1e. The points in the orange area in the ground truth map belong to class1, and the points in the blue area belong to class2. After training, prediction class map of each module was tested. Predicted class map of a module has some corrections on the basis of its previous module. The first four modules gradually approach the ground truth map, and the fifth module is degenerated into an identity mapping through shortcut connection, as the error of its previous module is very small .

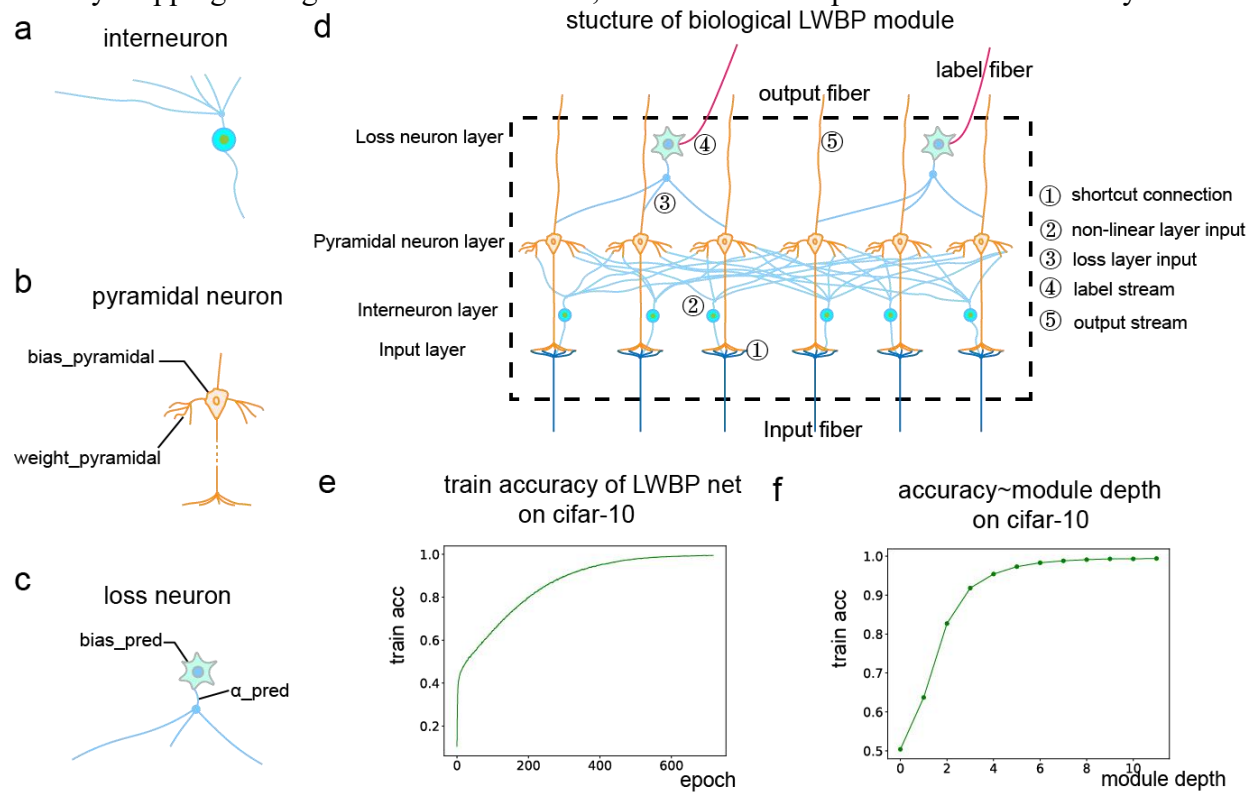

**Fig. 2| The neuron type, structure and performance of biological LWBP module.** Diagram of **a**, Interneuron, **b**, Pyramidal neuron, **c**, loss neuron. The modifiable parameters are marked with text. **d**, The structure of biological LWBP module. ①②③④⑤ corresponds to Fig. 1d. **e**, Train accuracy curve of biological LWBP network trained on cifar-10 data. **f,** Laye-wise accuracy of well trained biological LWBP network.

LWBP can theoretically be implemented in the form of biological neurons. We combined the structure of brain cortex to design a biological version of LWBP module. The biological LWBP module are composed of three kinds of neurons: interneuron, pyramidal neuron, and loss neuron, as shown in Fig. 2a-c. Parameters that can be updated have been marked with text, and the rest part are unchangeable. The structure of the entire module is shown in Figure 2d. Input fiber forms a shortcut connection with the apical dendrites of the pyramidal neurons. Interneurons receive input fiber and form full connections with somatic dendrites of pyramidal neurons. The output of the pyramidal neuron is transmitted to the next module, and its copy is passed to the loss neuron layer for calculating mean square error with label. To simplify the update formula, the connection between the loss neuron layer and the pyramidal neuron layer is set as a local connection, that a loss neuron occupies only a region of pyramidal neurons(See Methods for details). The biological version of LWBP module was also trained with cifar10. Training accuracy profile and every module's final accuracy are shown in Fig. 2e and Fig. 2f respectively. It needs to be pointed out that the biological LWBP module requires a large number of pyramidal neurons. Here 900 pyramidal neurons were used in each module to get a training accuracy of more than 99% in acceptable training epochs. BP algorithm only needs about 100 neurons per layer to get the same performance.

**Memory engram encoding algorithm and hippocampal fast memory**

According to memory engram theory, memories are believed to be encoded by sparse ensembles of neurons in the brain (*17*). We set the ideal engram encoding algorithm to have the following feature: when encoding any stimuli, 5% of neurons as the engram cells assigned for this stimuli will activate with value of 1. The rest 95% neurons will be inhibited. When encoding different stimuli, the activated neurons are different.

We designed a virtual experiment and a sparse coding method, for implement the ideal engram encoding algorithm. The experiment is shown as Fig. 3a, a virtual mouse walked randomly in a square box, it will meet a cheese at somewhere in the box. The sparse coding method is named Engram Autoencoder (EngramAE), which is a multi-layer autoencoder, as shown in Fig. 3b. EngramAE consists of three parts. The first part is an encoder composed of a multi-layer neural network, and its input is the coordinates (x, y). The second part is the engram neuron layer, which is composed of 1000 neurons with activation degree restricted to 0~1. The third part is the mapping matrix. the activation of engram neuron layer is multiplied with mapping matrix to get output ($x_{re}$, $y_{re}$).

EngramAE are trained with BP algorithm, with three parts of loss. The first loss is reconstruction loss, that is to make its output ($x_{re}$, $y_{re}$) equal to its input (x, y). In order to obtain sparse engram coding, the engram neuron layer is set to have the following sparse constraints:

1. When encoding a site, it is required that 5% neurons in the engram neuron layer have an activation degree of 1, and activation degree of the rest 95% neurons are 0. This constraint is to binarize the coding and regulate the proportion of activated neurons, witch will form engram coding.

2. The long-term average activation level of any engram neuron is limited to 5%. This constraint is to evenly distribute locations to all neurons, to prevent that some neurons from being overused, and some neurons have no effect.

We designed two sparse loss according to above sparse constraints. The overall loss function is the weighted sum of the three loss (see Methods for detail).

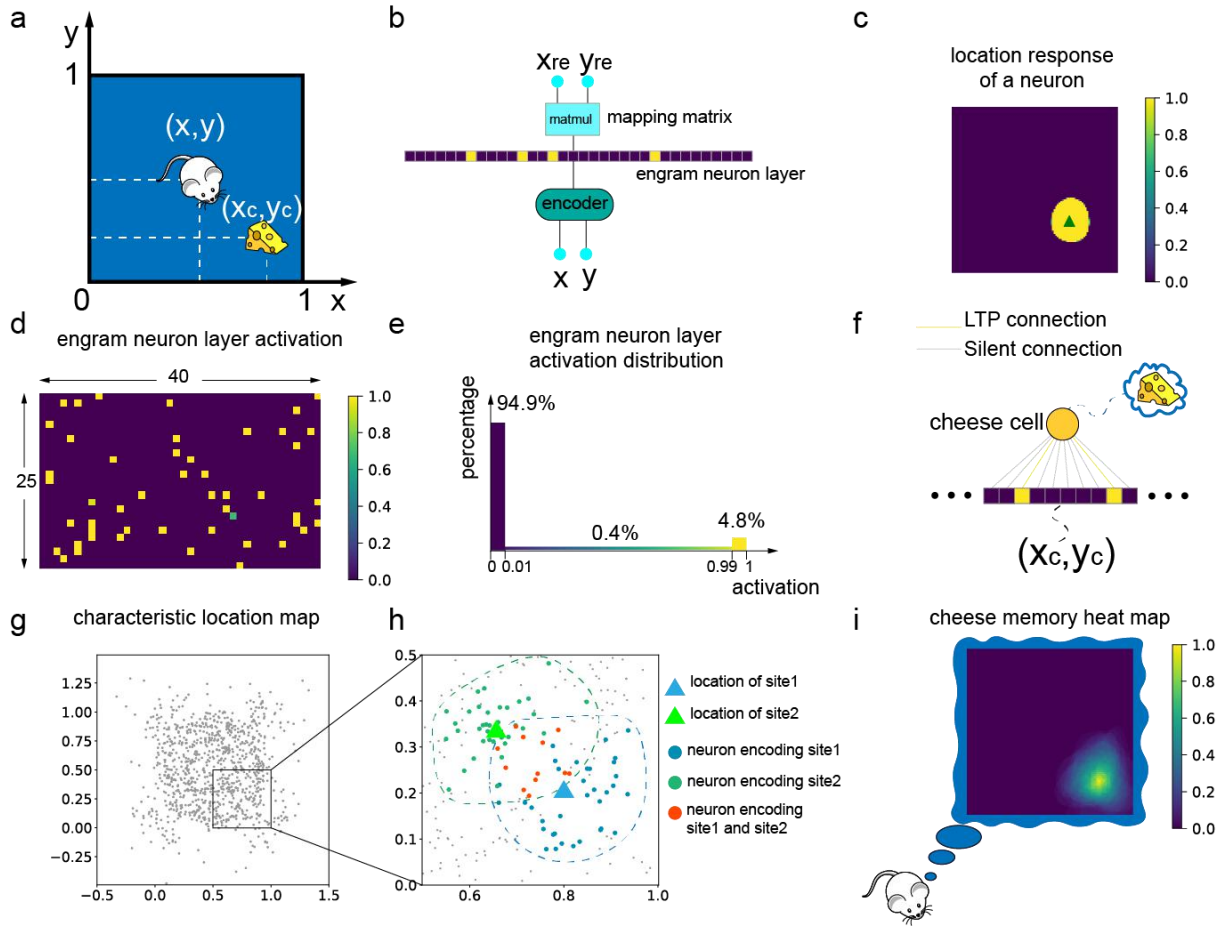

**Fig. 3| Engram Autoencoder completes positional associative memory tasks. a**, Diagram of virtual mice walking randomly in a box and meeting cheese in the corner. **b**, Engram Autoencoder consists of three parts: encoder, engram neuron layer, and mapping matrix. **c**, Location response of an engram neuron. The triangle symbol is the characteristic location of this engram neuron. **d**, Activation of all engram neurons at a certain location. 40×25=1000 is the total number of neurons **e**, Statistical results of the activation degree distribution of all engram neurons at over 10000 locations. **f**, The neuron encoding the cheese position (xc, yc) establishes LTP with the cheese neuron. **g**, Sactter plot of engram neurons' characteristic location. **h**, The triangle symbols represents site1 and site2. Neurons encoding site1 and site2 are highlighted by blue and green respectively, and the the neurons encoding both sites are highlighted by red. The dashed circle surrounds all neurons that encoding corresponding site. **i**, Heat map of the association memory between locations and cheese.

Input all positions during the randomly walking into the EngramAE for training. After trained, the engram neuron layer exhibits sparsity characteristics. If choose a neuron from engram neuron layer, and observe its response to the location, we will see the place field of this cell, as shown in Fig. 3c. A neuron only responds to a certain area, which is consistent with the behavior of place cell in hippocampus (*18*). The triangle symbol is the characteristic location of this neuron (see Methods). If choose a location and observe the activation of engram neuron layer, we will get engram coding, as shown in Figure 3d, that about 5% neurons are assigned to current location. If counting the activation of engram neuron layer at over 10000 locations, 94.9% of the neurons are inhibited(the activation value is less than 0.01), 0.4% of the neurons are activated between 0.01 and 0.99, and 4.8% of the neurons are extremely activated (activation value greater than 0.99). The proportion 4.8% is close to 5% that we set. The engram neuron layer shows binary engram coding, which is consistent with the memory engram theory. Two sparse constraints of the EngramAE are the driving force for the formation of engram cells.

When the mouse is at the location of cheese $(x_c, y_c)$, the engram neuron layer assigns 5% neurons for this location. Then these engram neurons will establish LTP connections with cheese neuron, which is the memory process of the hippocampus, as shown in Fig. 3f. If the mouse comes to $(x_c, y_c)$ in the future, it will trigger the memory of cheese through the established LTP connection. Will the location nearby $(x_c, y_c)$ trigger the memory of cheese? It depends on whether there are shared engram neurons between different locations. We draw the characteristic location of all neurons on a plane, as shown in Figure 3g (see Methods for drawing method). Input two different positions to EngramAE to get their assigned engram neurons and highlight them, as shown in Figure 3h. The neurons assigned only for site1 or site2 are highlighted by blue and green respectively, and the neurons encoding both sites are highlighted by red. Therefore, when the mouse is near $(x_c, y_c)$, it will trigger the memory of cheese in a weaker degree, through the LTP connection of shared engram neurons. The associative memory heat map about location and cheese is shown in Figure 3i.

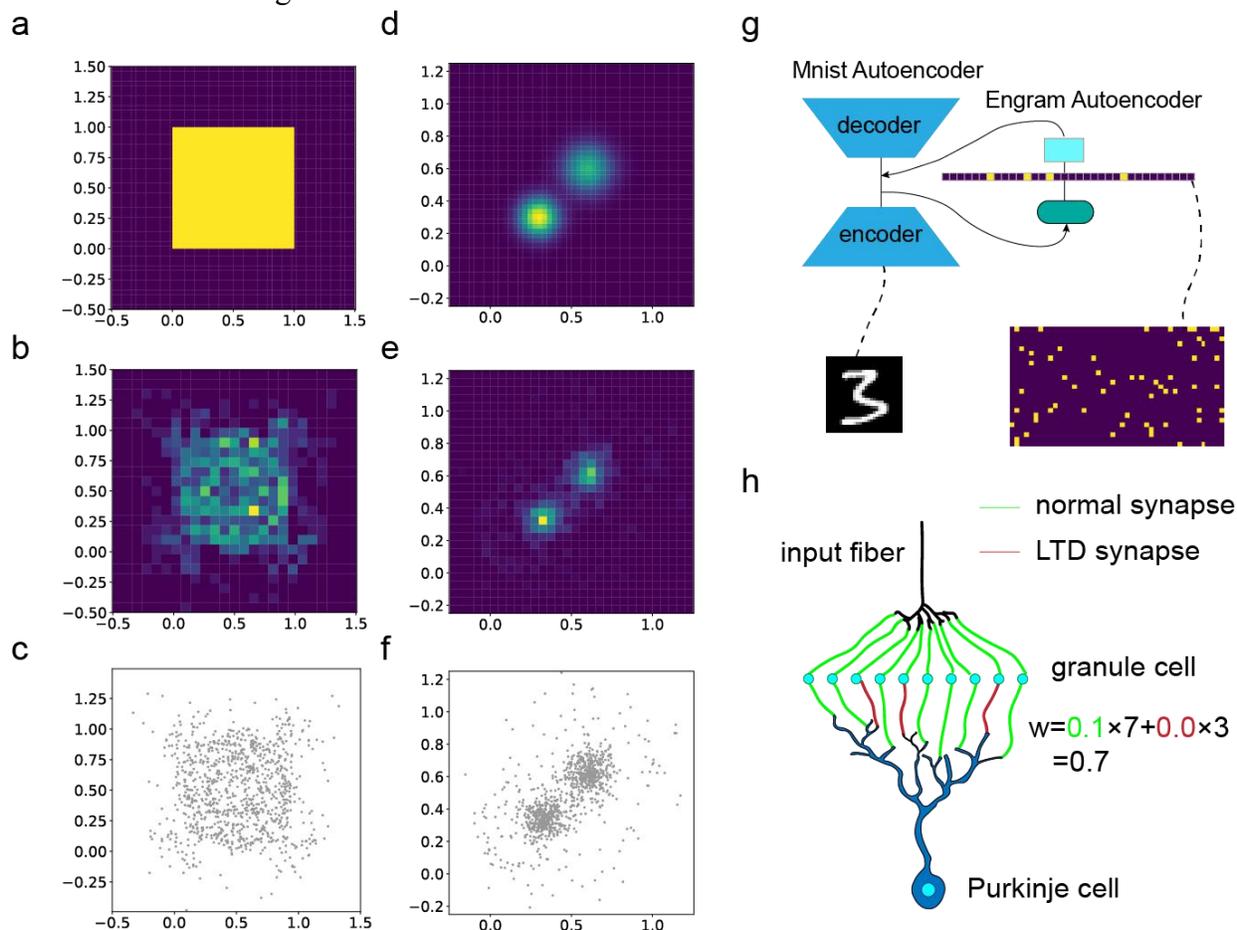

**Fig. 4| The performance of engram encoder under different data, diagram of cerebellar effective weight adjustment. a**, The position distribution of random walking in a box is equivalent to square uniform distribution. **b**, Density heatmap and **c**, scatter plot of characteristic

location of all neurons in engram neuron layer, after trained with square uniform distribution. **d**, A bimodal gaussian distribution. **e**, Density heatmap and **f**, scatter plot of characteristic location of all neurons in engram neuron layer, after trained with bimodal gaussian distribution. **g**, The dimension of Mnist data is reduced by multi-layer autoencoder and then input into Engram Autoencoder for sparse encoding. **h**, 10 granular cells are used to disperse the effective weight between Purkinje cells and the input fiber, to achieve rapid adjustment of the effective weight. The sum of the effective weights formed by 10 granular cells is 1, which becomes 0.7 after three synapses are inhibited by LTD .

The above result shows that each engram neuron has a characteristic location and place field. Distribution of engram neuron's characteristic location are related to the distribution of real potion. The mouse walked randomly in the box, and its real position distribution is a square uniform distribution, as shown in Fig. 4a. By drawing the characteristic location of all 1000 neurons, we can get a scatter plot, as shown in Fig. 4c. The density heat map of characteristic location point is shown as Fig. 4b, which is close to the real location distribution. This means that neurons are evenly distributed to the location space. Further, we tested the gaussian position distribution. In Fig. 4d-f, if train EngramAE with the bimodal gaussian distribution, the characteristic location distribution are close to the bimodal gaussian distribution. It could be concluded that areas with high frequency are assigned more neurons. EngramAE could also encode complex data. More general data distribution such as Mnist image can also be encoded with EngramAE, and each picture is also assigned with an ensemble of engram cells, as shown in Figure 4g. But the dim of the image need to be reduced through a traditional auto-encoder (see Methods for details). The results show that the engramAE can sparsely encode a variety of data, so it is a universal encoding method.

**Cerebellar LTP/LTD and rapid learning**

The cerebellum needs to complete learning tasks about continuous value , such as decreasing the strength of a muscle by 30%. Here the role of LTP/LTD is different from that of hippocampal memory.We proposed a mechanism for the cerebellum to achieve rapid learning with LTP/LTD, as described below. Purkinje cells are the main processing neurons in the cerebellum, and there are a lot of granule cells connected to it. Assume that 10 granular neurons receive input and pass information to Purkinje neurons, each connection with Purkinje cells has a weight of 0.1, and the effective weight from input fibers to Purkinje cells is 1, as shown in Fig. 4h. The LTD could be used to quickly adjust the effective weight. To achieve the effective weight of 0.7, the cerebellum only needs to inhibit 3 connections by LTD. Similarly, if the goal is to increase the effective weight, the cerebellum can use the LTP mechanism to activate some connections.

**Discussion**

LWBP module can update the parameters immediately after calculating the loss by itself, while BP needs to wait for the return of error gradient from the end of network. The weight update of every LWBP module runs independently, which is suitable for multi-processor distributed computing, and the cerebral cortex is a kind of distributed computing. The disadvantage of LWBP algorithm is that training is 10 times slower than BP and requires a large number of neurons. Maybe this is the reason why there are billions of neurons in the cerebral cortex. LWBP experiment did not consider the convolutional layer in artificial intelligence, because similar functional areas in the brain exist between the retina and the visual cortex, the function is generated congenitally rather than learned (*19*). Cortical connections are in the form of fully connected layers. However, LWBP is a supervised learning algorithm that could make use of deep network, which is also biologically deployable. We think this may be the closest algorithm to the cerebral cortex. Whether LWBP is the learning algorithm of the cerebral cortex still needs biological experimental verification.

In the EngramAE experiment, density of engram neurons' characteristic location distribution is proportional to the density of data distribution, which indicates that the EngramAE will automatically allocate more neurons for more frequently occurring data. This property of EngramAE is coincident with some biological phenomena. For example, people use hands frequently, so the area of the sensory cortex related to the hands are large (*20*), which reflects the adaptive allocation function of brain. EngramAE could encode high-dimensional data like image. It could be considered that EngramAE is a universal form of coding, which not only participates in location memory, but also could be used in high-level cognitive memory. We believe that the sparse constraint of EngramAE exist in the brain, which requires experimental verification.

Cerebellum uses LTP/LTD to fast adjust the effective weight from input fiber to Purkinje neuron. To complete the same tasks, the cerebral cortex has to rely on slow synaptic chemical changes, which will take a long time. If there are 100 granular neurons connected to Purkinje neuron, the adjustment precision will change from 0.1 to 0.01, if the total weight is still 1. The precision of effective weight adjustment is proportional to the number of granule cells that connected to Purkinje cell. The ratio of number between granule cells to Purkinje cells in the human cerebellum is 2991:1, which is much higher than that of other mammals (*21*). It could be inferred with our theory that the human cerebellum can complete detailed and fast learning tasks.

**Acknowledgments:**

We thank Li Yi for teaching figure drawing. **Funding:** None. **Author contributions:** Yifei Mao do experiments and write paper; **Competing interests:** The authors declare no competing interests.

**Data and materials availability:** All data are available in the main text or the supplementary materials. The computer code are available at https://github.com/maofeifei94/Brain_Learning_Memory.

# Materials and Methods

## Devices and tools

In all computation experiments, operating system is Windows 10 and Ubuntu 16.0, cpu is intel i5 6500, gpu is nvidia GTX 1070. The programming language is python3.6, algorithm about neural networks are implemented with pytorch1.6. All codes are available at https://github.com/maofeifei94/Brain_Learning_Memory.

## Predefined operation

The formula for the linear layer is:
$$y = x \cdot W + b$$

Where x is input with dimension of [in_dim], y is output with dimension of [out_dim], W is a learnable matrix with shape of [in_dim,out_dim], the operation between W and x is matrix multiplication. b is biast with dimension of [out_dim].

The operation of the nonlinear layer is:
$$y = actfunc(x \cdot W + b)$$

Where actfunc is a non-linear function. Tanh, Sigmoid, LeakyRelu are used in this article.

If it is a nonlinear layer with shortcut connection:
$$y = actfunc(x \cdot W + b) + x$$

A loss layer consists of a nonlinear layer and loss function:
$$Loss = lossfunc(actfunc(xW + b), label)$$

Where W transform the Matrix. W converts the dimension of input into the dimension of label. The loss function used in this article are cross entropy loss and mean square error.

## Train network with LWBP and BP on cifar-10

We tested the LWBP and BP algorithm on cifar-10 dataset. Cifar-10 is an image classification dataset containing 10 types of pictures such as airplane, bird, cat, etc. The picture resolution of cifar-10 is [32,32], there are 5000 pictures of each category, a total of 50,000 training pictures. The picture is normalized before being input to the network for training:
$$x_{norm} = \frac{x - mean(x)}{stddev(x)}$$

The optimization method in each module of LWBP is backpropagation. Data is forwarded between modules, and gradients are limited within the module with the detach function of pytorch. Three comparative experiments were done on cifar-10 data: LWBP with shortcut connection, LWBP without shortcut connection, and BP with shortcut connection.

The following are the common characteristics of the three experimental networks: network of the three experiments are composed of 12 layers, the dim of each layer is 300, and the activation functions are all Tanh. The optimizer is Rmsprop, learning rate is 0.0001, batchsize is 256, and the loss functions are all cross-entropy. Data dimension of cifar-10 is 32×32×3=3072. It should be noted that the when the first module (or layer) compresses the data dimension from 3072 to 300, the input and output dimensions are inconsistent. Therefore, the first module (or layer) of the three experiments did not use shortcut connections, and the first module (or layer) did not use the activation function to better demonstrate the ability of LWBP to use the deep network.

There are some differences between the three experiments: The network of BP algorithm experiment has a shortcut connection, and there is a linear layer added at the end of the network for prediction and loss calculation. The network of the LWBP algorithm experiment uses a linear layer for prediction and loss calculation in each LWBP module. Top1 criterion is used to calculate the prediction accuracy, that is, to determine whether the category with the highest predicted probability is equal to label.

**Train network with LWBP on 2d point classification task**

Here LWBP algorithm was tested with 2d point data.Data are points on a 2d plane, and the range of x and y are both 0 to 1. The dividing line of two class is an inclined sine function, as shown in the Fig. s1a, angle between x' axis and x axis is 45 degrees, and the dividing line function is:

$$y' = 0.4\sin(4 \cdot \pi \cdot \frac{x'}{\sqrt{2}})$$

Points in orange area are labeled 0, blue area are labeled 1. 2d points in the rectangular area are randomly selected and input into the network, and their labels are transferred to each module. The network structure of this experiment is composed of 5 LWBP module with shortcut connection. Dimension of each layer is 16 neurons. The first module did not have shortcut connection and not use activation function, the reason is the same to previous section. Activation function of the nonlinear layer is LeakyRelu, and loss function is cross entropy. Batchsize is 256, optimizer is RMSprop, learning rate is 0.0001. While test, 150×150 grid points are taken at equal intervals from the entire rectangle area, and input to the network to obtain a predicted heat map of each module with resolution of 150×150 .

**Biological version of LWBP**

The biological version of the LWBP module consists of three types of neurons, namely interneurons, pyramidal neurons, and loss neurons. Interneurons and input fibers are connected in one-to-one correspondence. Interneurons only play the role of spreading information, their synapse weights are all 1, and cannot be modified. The parameter of pyramidal neurons and loss neurons is shown in Fig. s1b, c. The structure of the LWBP module is shown in Fig. s1d. Because cross-entropy loss is difficult to deploy using biological neurons, the loss function here is the mean square error loss. The calculation formula from Input to Output is:

$$Output = Input + Tanh(Input \cdot W_{pyramidal} + b_{pyramidal})$$

Here $W_{pyramidal}$ is the connection matrix of pyramidal neuron body dendrites and interneurons, $b_{pyramidal}$ is the pyramidal neuron bias, as shown in Fig. s1b. The first term of the above formula comes from the shortcut connection between the apical dendrites and the Input, and the second term comes from the connection between the somatic dendrites and the interneuron. For simplifying the synapse update formula, the connection between the loss neuron layer and the pyramidal neuron layer is set local connection. Each loss neuron connects only a region of pyramidal neurons, and performs an average operation on these neurons. At the same time, each pyramidal neuron belongs to only one loss neuron. In this way, each pyramidal neuron only needs to consider to the corresponding loss neuron when updating the weight, and doesn't need to consider all the loss neurons like what fully connected layer does. In this experiment, because the dimension of the cifar-10 label is 10, so the number of loss neurons is 10. The output dimension is 900, so each loss neuron receives the output of 900/10=90 pyramidal neurons. The formula for calculating the mean value of pyramidal neurons that connected to k-th loss neuron is as follows:

$$OutMean_k = \frac{\sum_{j=k \cdot \rho}^{(k+1) \cdot \rho} Output_j}{\rho}$$

Where k∈[1,10] is the index of the loss neuron, j∈[kρ,(k+1)ρ) is the index of the pyramidal neuron, and ρ=90 is the number of pyramidal neurons that received by each loss neuron. The calculation formula for prediction and loss is:

$$pred_k = Sigmoid(\alpha_{pred\_k} \cdot OutMean_k + b_{pred\_k})$$

$$Loss = \frac{\sum_{k=1}^{10}(Label_k - pred_k)^2}{10}$$

Where $pred_k$ is the prediction inside the k-th loss neuron (not explicitly output), $\alpha_{pred\_k}$ is the scaling factor of the k-th loss neuron to OutMean, and $b_{pred\_k}$ is the bias of the k-th loss neuron. The whole module has 4 learnable parameters. $W_{pyramidal}$ and $b_{pyramidal}$ are the learnable parameters of pyramidal neurons, $\alpha_{pred}$ and $b_{pred}$ are the learnable parameters of Loss neurons. The update formula of the four parameters is:

$$\Delta\alpha_{pred\_k} = -lr \cdot \frac{\partial Loss}{\partial pred_k} \cdot \frac{\partial pred_k}{\partial \alpha_{pred\_k}}$$

$$= lr \cdot 2 \cdot \frac{(Label_k - pred_k)}{10} \cdot pred_k \cdot (1 - pred_k) \cdot OutMean_k$$

$$\Delta b_{pred\_k} = -lr \cdot \frac{\partial Loss}{\partial pred_k} \cdot \frac{\partial pred_k}{\partial b_{pred\_k}}$$

$$= lr \cdot 2 \cdot \frac{(Label_k - pred_k)}{10} \cdot pred_k \cdot (1 - pred_k)$$

$$\Delta W_{pyramidal\_ji} = -lr \cdot \frac{\partial Loss}{\partial pred_k} \cdot \frac{\partial pred_k}{\partial OutMean_k} \cdot \frac{\partial OutMean_k}{\partial Output_j} \cdot \frac{\partial Output_j}{\partial W_{pyramidal\_ji}}$$

$$= lr \cdot 2 \cdot \frac{(Label_k - pred_k)}{10} \cdot pred_k \cdot (1 - pred_k) \cdot \alpha_{pred\_k} \cdot \frac{1}{\rho} \cdot (1 - Output_j^2) \cdot Input_i$$

$$\Delta b_{pyramidal\_j} = -lr \cdot \frac{\partial Loss}{\partial pred_k} \cdot \frac{\partial pred_k}{\partial OutMean_k} \cdot \frac{\partial OutMean_k}{\partial Output_j} \cdot \frac{\partial Output_j}{\partial b_{pyramidal\_ji}}$$

$$= lr \cdot 2 \cdot \frac{(Label_k - pred_k)}{10} \cdot pred_k \cdot (1 - pred_k) \cdot \alpha_{pred\_k} \cdot \frac{1}{\rho} \cdot (1 - Output_j^2)$$

Where k∈[1,10] is the index of the loss neuron, j∈[kρ,(k+1)ρ) is the index of the pyramidal neuron, and i is the index ∈[1,900] of the input fiber and interneuron. lr is the learning rate. $W_{pyramidal\_ji}$ represents the synapse weight between the j-th pyramidal neuron and the i-th interneuron. All the information needed to update the weight of the neuron can be found within the module, and the gradient information of a pyramidal neuron only depends on the corresponding loss neuron, so the gradient information may be transmitted through the neurotransmitter or the feedback connection from loss neuron.

**Engram Autoencoder for location associative memory**

In this experiment, the mouse is simplified into a single point which moves randomly in a box of 1×1 size. The direction of movement is random, with a step length of 0.02．the location of cheese (xc,yc)=(0.8,0.2).The position coordinates of the mouse are input into EngramAE. The detailed structure of the Engram Autoencoder is shown in Fig. s2. The encoder is a nonlinear multi-layer network, and the size of each layer is [64,64,256,256,256,256]. The activation function of each layer is LeakyRelu. The engram neuron layer has a total of 1000 neurons, which are fully connected with the encoder, and the activation function is Sigmoid to constrain the output to 0~1. The size of the mapping matrix is [1000, 2], it performs matrix multiplication with the Engram neuron layer, and the two-dimensional vector in each row represents the characteristic location of a neuron. Engram Autoencoder is optimized by BP algorithm, the optimization method is Rmsprop, and the learning rate is 0.0001. The overall loss consists of reconstruction loss, engram sparse loss, and long-term activation loss. The reconstruction loss $L_{reconstruction}$ is mean square error calculated as follows:

$$L_{reconstruction} = \frac{(x-x_{re})^2 + (y-y_{re})^2}{2}$$

$x_{re}$ and $y_{re}$ are the reconstructed output values of the Engram Autoencoder.

When encoding a site, it is required that 5% of the engram neurons has an activation degree of 1, and activation degree of the rest 95% neurons are 0. The corresponding loss is engram sparse loss $L_{engram\_sparse}$, which is calculated as follows:

$$L_{engram\_sparse} = \left| \sum_{i=1}^{1000} h_i^2 - \cdot 5\% \right| + \left| \sum_{i=1}^{1000} (1-h_i)^2 \cdot 1000 \cdot (1-5\%) \right|$$

Where $h_i$ is the activation value of the i-th neuron in the engram neuron layer, and η=5% is the sparsity coefficient.

$L_{time\_sparse}$ constrains the long time average activation degree of neurons in the engram layer. The purpose is to make an engram cell have a strong response to 5% of the data and inhibit the response to the rest. Exponential moving average was used to count the long-term average activation of neurons:

$$avg\_h_t = avg\_h_{t-1} \times \lambda + (1-\lambda) \times h$$

Where λ is the discount factor, $h_i$ is the activation value of the i-th neuron in the engram neuron layer, and avg_h is their moving average. If λ=0.9999, the inertia of the system is too strong, which may cause large fluctuations. So we adopted a PID-like control method and calculated long time average activation long_avg_h with λ=0.9999, and short time average activation short_avg_h with λ=0.99.

$$long\_avg\_h_t = long\_avg\_h_{t-1} \times 0.9999 + (1-0.9999) \times h$$
$$short\_avg\_h_t = short\_avg\_h_{t-1} \times 0.99 + (1-0.99) \times h$$

The calculation formula of $L_{time\_sparse}$ is:

$$L_{time\_sparse} = \frac{1}{1000} \cdot \sum_{i=1}^{1000} (-\frac{5\%}{avg\_h_{t\_i}} + \frac{1-5\%}{1-avg\_h_{t\_i}}) \times h_i$$

The overall long-term average activation sparsity loss is:

$$L_{longshort\_time\_sparse} = 0.9 \cdot L_{time\_sparse} |_{\lambda=0.9999} + 0.1 \cdot L_{time\_sparse} |_{\lambda=0.99}$$

The overall loss of the Engram Autoencoder is

$$Loss_{EngramAE} = k_1 \cdot L_{reconstruction} + k_2 \cdot L_{engram\_sparse} + k_3 \cdot L_{longshort\_time\_sparse}$$

k1, k2, k3 are 1000, 0.01, and 10 respectively.

After the training is completed, we take grid points of [101x101] at equal intervals from the box area, and input these points into the Engram Autoencoder. Take the activation degree of an engram neuron at each grid point to make the position response map of the engram neuron, shown as Fig. s4a．Take the activation value of all engram neurons in a position to get the engram sparse coding of this position, shown as Fig. s4b. If we statistic the activation of all neurons at all grid points (101×101), we could get the activation distribution of the engram neuron layer.

The drawing method of characteristic location scatter plot is shown in Fig. s2b, c, that taking a row vector from the mapping matrix we can get the characteristic location of a neuron. Draw each characteristic location on a 2d plane to obtain the characteristic location scatter plot of all neurons, as shown in Fig. s2d. The density heat map are drawing with matplotlib of python.

When in the position of cheese, the weight of the connection between the engram neuron with activation value greater than 0.95 and the cheese neuron is set to 1, indicating that LTP is formed. The connection weight between the rest engram neurons and cheese neurons is set to 0. Input 101×101 grid points to the Engram Autoencoder, and calculate the cheese memory degree of each location, we will get a memory heat map about cheese.

**Engram Autoencoder and bimodal Gaussian distribution**

When training Engram Autoencoder, sample from two Gaussian distributions with equal probability to form a batch of data. The parameter for two Gaussian distributions is

$$\mu_1 = (0.3, 0.3), \Sigma_1 = \begin{pmatrix} 0.08 & 0 \\ 0 & 0.08 \end{pmatrix}$$

$$\mu_2 = (0.6, 0.6), \Sigma_2 = \begin{pmatrix} 0.1 & 0 \\ 0 & 0.1 \end{pmatrix}$$

Where $\mu_1$, $\mu_2$ are the center of two Gaussian distributions, and $\Sigma_1$, $\Sigma_2$ are covariance matrices. The structure, parameters and loss function of Engram Autoencoder are the same to previous section.

**Engram Autoencoder and Mnist data**

The dimension of the Mnist picture is 28×28=784. Engram encoding of Mnist requires dimension reduction of the data. We use traditional multi-layer autoencoders for dimension reduction, the autoencoder used for Mnist dimension reduction is named Mnist Autoencoder (MnistAE). The structure of the entire network is shown in Fig. s3. The dimension of each layer of the MnistAE's encoder is [256,256,256,128]. MnistAE's encoder reduces the dimension of the Mnist picture to 128 dimensions and then inputs it into the Engram Autoencoder for engram encoding. The input vector and output vector of the Engram Autoencoder are averaged and then input into the decoder of the Mnist Autoencoder. The dimension of each layer of MnistAE's decoder is [256,256,256,784]. The activation function of the layer marked in Fig. s3 is Sigmoid, and the activation of the rest layers are LeakyRelu.

The reconstruction loss of MnistAE is the mean square error loss, and its calculation formula is

$$\text{Loss}_{MnistAE} = \frac{\sum_{i=1}^{784}(img_i - img_{re\_i})^2}{784}$$

The overall loss is the weighted sum of the reconstruction loss of MnistAE and the loss of Engram Autoencoder:

$$\text{Loss}_{all} = 100 \cdot \text{Loss}_{MnistAE} + \text{Loss}_{EngramAE}$$

The optimization method is Rmsprop, and the learning rate is 0.0001. Fig. s4c shows the input image, resconstructed image and engram coding of well trained MnistAE.

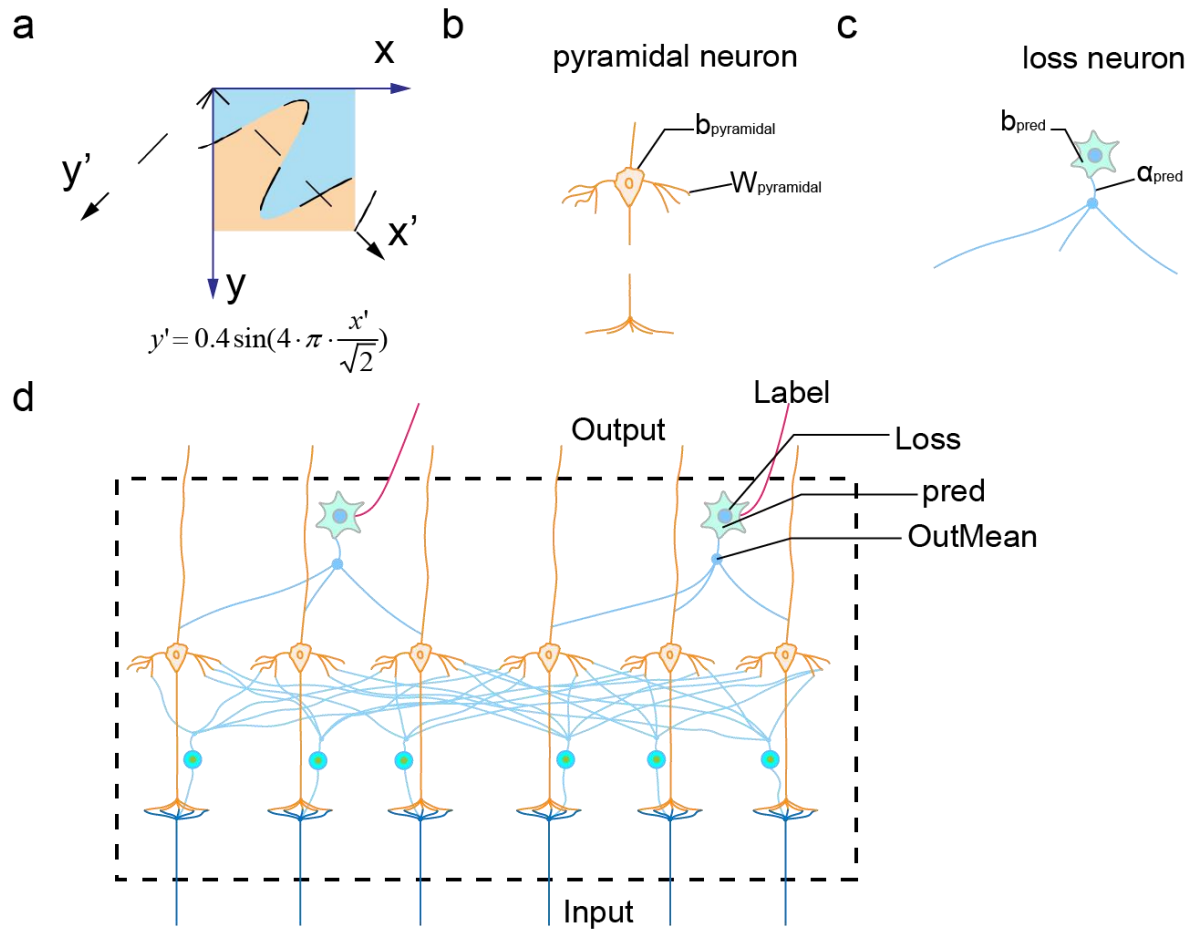

**Fig. S1. Diagrams related to LWBP experiment. a**, The function of the dividing line of two-dimensional point classification task. **b**, Pyramidal neurons have two learnable parameters, which are synaptic weight of somatic dendrites and bias. The synaptic weight of Apical dendrite is 1, and unmodifiable. **c**, The synaptic weight of loss neuron's dendrites is Equal to 1 divided by the number of connected pyramidal neuron, and unmodifiable. Loss neurons have two learnable parameters, which are scaling factor α and bias. **d**, Some nodes in the calculation process. The outmean node represents the loss neuron to average the received signal. The calculation of pred and Loss is done inside the loss neuron cell body.

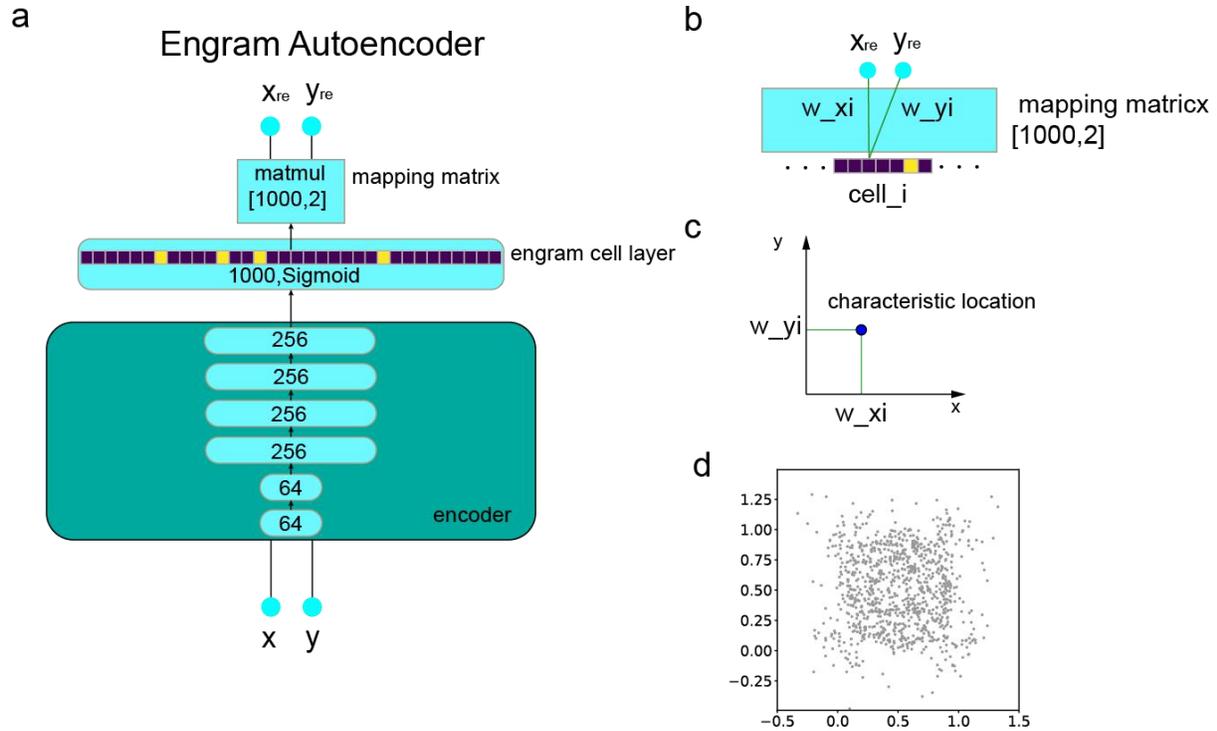

**Fig. S2. The structure of Engram Autoencoder and the drawing method of characteristic location scatter plot. a**, Structure of Engram Autoencoder. The encoder is a nonlinear multi-layer network, the activation function of each layer is LeakyRelu. The engram neuron layer has 1000 neurons,which are fully connected with the encoder, its activation function is Sigmoid to constrain the output to 0~1. The size of the mapping matrix is [1000, 2], it performs matrix multiplication with the Engram neuron layer. **b**, Each row of the mapping matrix represents the characteristic location of a neuron, which is (w_xi,w_yi) for the i-th neuron. **c**, Draw (w_xi,w_yi) in the coordinate system. **d**, Draw (w_xi,w_yi) of all 1000 neurons in the coordinate system, we will get the characteristic location scatter plot.

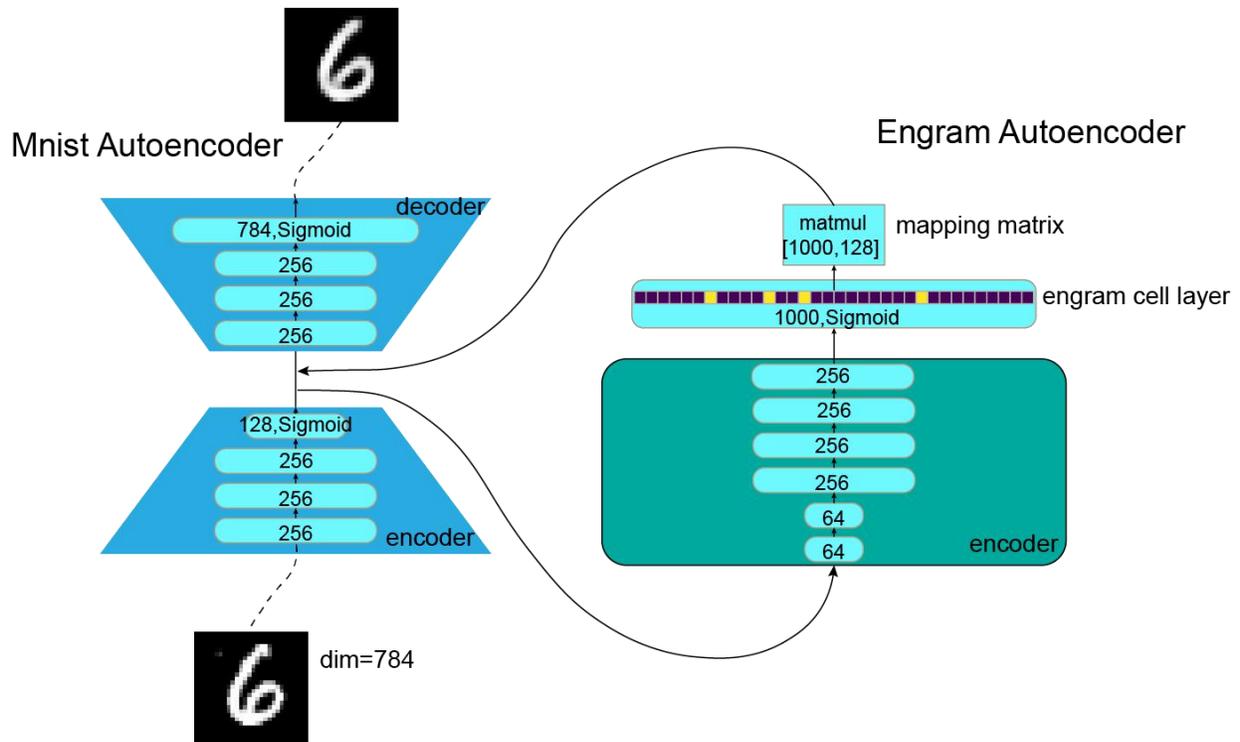

**Fig. S3. The structure of Mnist Autoencoder and Engram Autoencoder.** Mnist Autoencoder is a traditional multi-layer autoencoders for dimension reduction. The dimension of data input to Engram Autoencoder is 128. Except for the layers marked with Sigmoid, the activation function of other layers is LeakyRelu. The input vector and output vector of the Engram Autoencoder are averaged and then input into the decoder of the Mnist Autoencoder. The whole network are optimized by Rmsprop.

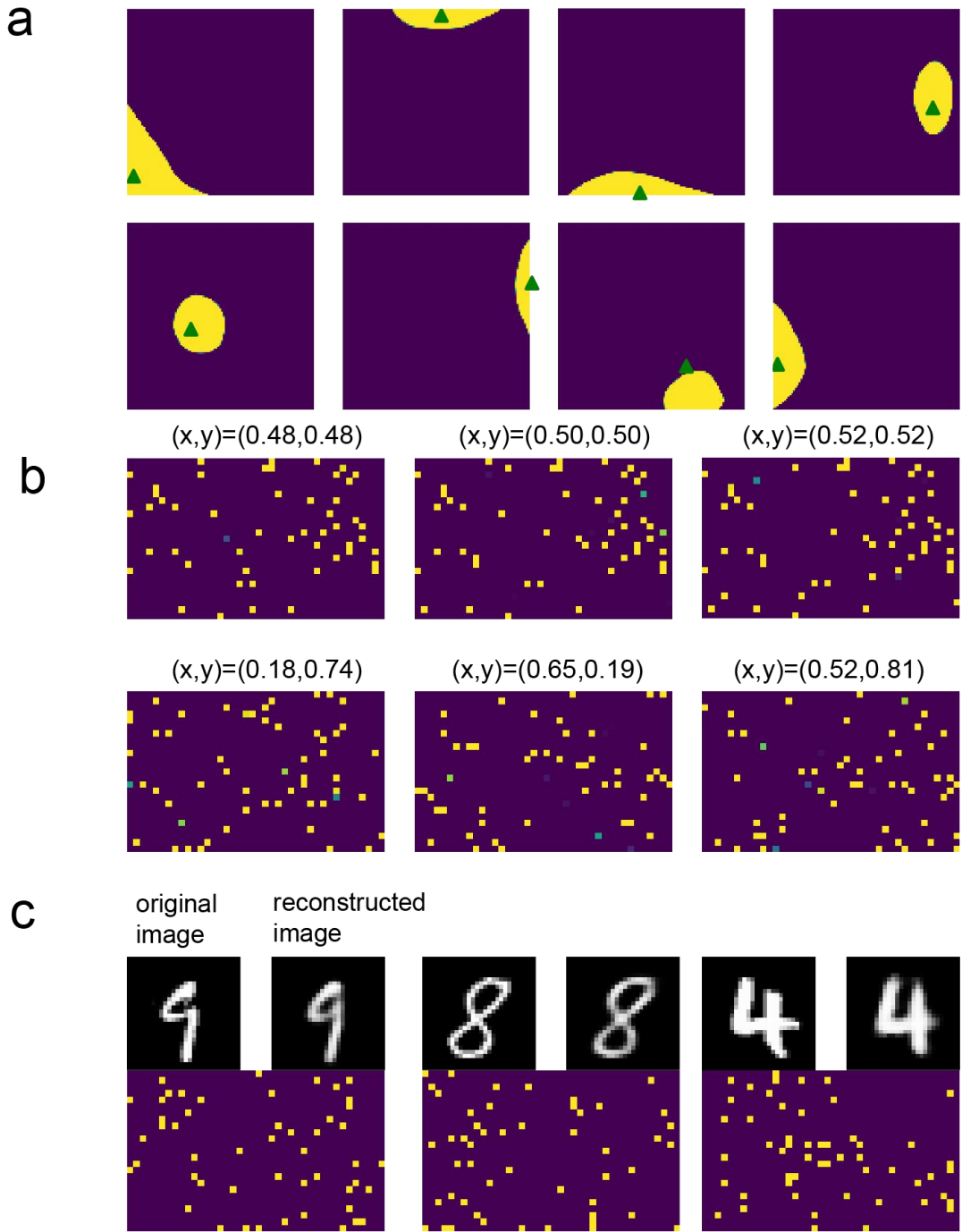

**Fig. S4. Some coding legends of Engram Autoencoder. a**, Location response of some neurons from engram neuron layer. The data input to Engram Autoencoder is the location where the mouse walks randomly in the square box. The triangle symbol is the characteristic location of this neuron. **b**, Engram neuron layer population activation in some locations. The site (x,y) of the top three pictures are very close, so their coding are only slightly different. **c**. The original image, reconstructed image of Mnist Auencoder, and the corresponding Engram neuron layer population activation map.